\documentclass[11pt]{article}
\usepackage[final]{acl}

\usepackage{times}
\usepackage{latexsym}
\usepackage[T1]{fontenc}
\usepackage[utf8]{inputenc}
\usepackage{microtype}
\usepackage{inconsolata}
\usepackage{graphicx}
\usepackage{booktabs}
\usepackage{enumitem}
\usepackage{url}
\usepackage{hyperref}
\usepackage{xspace}
\usepackage{tabularx}

\newcommand{\system}{\textsc{Oracle}\xspace}
\newcommand{\ours}{Time-Dependent Recursive Summary Graph (TRSG)\xspace}
\newcommand{\pestel}{PESTEL\xspace}

\title{\system: Time-Dependent Recursive Summary Graphs for Foresight\\
on News Data Using LLMs}

\author{Lev Kharlashkin, Eiaki Morooka, Yehor Tereshchenko and Mika Hämäläinen \\
  Metropolia University of Applied Sciences \\
  Helsinki, Finland \\
  \texttt{first.last@metropolia.fi} \\}

\begin{document}
\maketitle

\begin{abstract}
\system\ turns daily news into week-over-week, decision-ready insights for one of the Finnish University of Applied Sciences. The platform crawls and versions news, applies University-specific relevance filtering, embeds content, classifies items into \pestel\ dimensions and builds a concise \emph{\ours}: two clustering layers summarized by an LLM and recomputed weekly. A lightweight change detector highlights what is \emph{new}, \emph{removed} or \emph{changed}, then groups differences into themes for \pestel-aware analysis. We detail the pipeline, discuss concrete design choices that make the system stable in production and present a curriculum-intelligence use case with an evaluation plan.
\end{abstract}

\section{Introduction}

Foresight is the systematic detection, interpretation, and anticipation of signals of change to inform long-term decision-making \citep{chandrasekaran2022future,yuksel2012pestel}. 
For our setting, this means turning large, unstructured text streams (news, reports, social media) into structured representations of emerging trends and weak signals that remain interpretable to human analysts. 
Public institutions and universities in particular need early indicators of technological, policy, or societal change to align curricula, partnerships, and research priorities \citep{hamalainen2024suositus}. 
Traditional approaches (expert workshops and static reports) cannot match the velocity and volume of modern information flows, motivating continuously operating, evidence-traceable foresight systems.

\system\ addresses this need for a University of Applied Sciences by transforming Finnish news into a weekly \emph{recursive summary graph} that reveals how narratives emerge, merge, and fade over time, grounded in \pestel\ dimensions (see \citealt{yusop2018pestel}). Unlike generic dashboards, \system\ is designed for ongoing operation, traceability, and institutional relevance.

Our contributions are:
\begin{itemize}[leftmargin=1.05em,itemsep=2pt,topsep=2pt]
    \item A practical pipeline for daily Finnish news ingestion with hashing-based versioning and vectorized storage.
    \item A two-level \ours\ that hierarchically organizes narratives and updates weekly while accumulating long-term knowledge.
    \item A week-to-week change-detection mechanism that groups themes and analyzes institutional relevance through \pestel.
    \item A real-world use case—curriculum intelligence—demonstrating decision support for academic stakeholders.
\end{itemize}

\section{Related Work}

\paragraph{Semantic representations and retrieval.}
Dense sentence embeddings (e.g., SBERT, SimCSE) enable semantic clustering and temporal tracking over large text streams \citep{reimers2019sentence,gao2021simcse}. For scalable retrieval, approximate nearest-neighbor search and vector databases such as FAISS and Milvus support efficient similarity queries over continuously ingested corpora \citep{johnson2019billion,milvus2021}.

\paragraph{Clustering and community detection.}
Graph-based community detection (Louvain, Leiden) is widely used to expose latent structure in text-derived graphs \citep{blondel2008louvain,traag2019louvain}. Hybrid topic models like BERTopic combine transformer embeddings with clustering to yield interpretable topics, though they are typically applied to static snapshots rather than rolling streams \citep{grootendorst2022bertopic}.

\paragraph{Summarization and abstraction.}
Neural abstractive models (BART, PEGASUS) produce fluent summaries but face challenges in factual grounding at multi-document scale \citep{lewis2020bart,zhang2020pegasus}. Extractive methods (LexRank, TextRank) remain strong for faithfulness and traceability \citep{erkan2004lexrank,mihalcea2004textrank}. Streaming and dynamic summarization explores hierarchical, time-aware abstractions over evolving corpora \citep{huang2024streamsum}.

\paragraph{Temporal modeling and change detection.}
Modeling evolving topics spans burst detection and dynamic topic models, capturing intensity and drift over time \citep{kleinberg2003bursty,blei2006dynamic}. These lines inform explicit week-over-week change labeling and rolling updates for interpretable monitoring.

\paragraph{LLMs in foresight and strategic analysis.}
Recent work investigates LLMs for structured foresight, including multi-layer \pestel-based prompting to scaffold analysis \citep{alnajjar2024mlpestel}. Our system operationalizes these ideas as a continuous, interpretable pipeline combining semantic retrieval, dynamic clustering, hierarchical summarization, and explicit change tracking tailored to institutional decision-making.

\section{Oracle Platform}
\begin{figure*}[t]
  \centering
  \includegraphics[width=\textwidth]{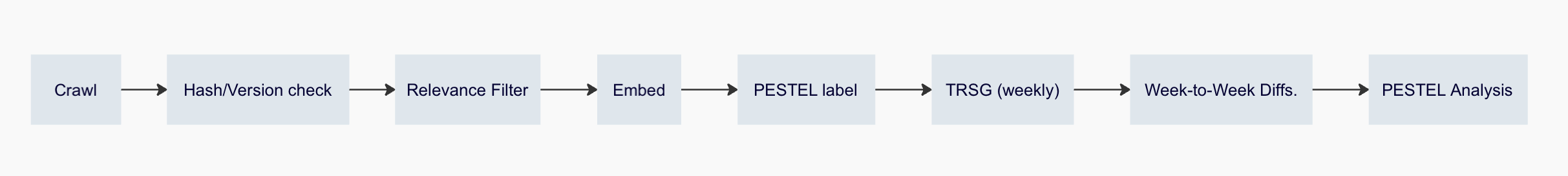}
  \caption{\system\ workflow overview.}
  \label{fig:pipeline}
\end{figure*}

\subsection{Data Ingestion and Versioning}
\paragraph{Sources.} Finnish news (e.g., Yle) using open-source RSS feeds are crawled daily. The pipeline extracts canonical URLs and main content (boilerplate-stripped), preserving the original HTML for audit.

\paragraph{Hashing.} We compute a stable content hash over normalized HTML. If the hash changes for a known URL, a new version is stored and re-embedded. This suppresses duplicates while tracking edits (e.g., headline updates).

\subsection{Relevance Filtering for University}
\label{sec:relevance}
A two-stage filter prioritizes items that matter institutionally:
\begin{enumerate}[leftmargin=1.1em,itemsep=1pt,topsep=2pt]
    \item \textbf{Lexical stage.} Query expansions over names (\emph{University}, \emph{UAS}, Finnish aliases), domains (education, R\&D, local industry) and geography. This fast pass removes obviously unrelated stories.
    \item \textbf{Semantic stage.} Embedding-based similarity against curated exemplars (e.g., skills funding, curriculum reform, regional innovation). Borderline items are kept if semantically close.
\end{enumerate}
Non-relevant items are cold-stored for later replays (interests can shift).

\subsection{Embeddings, Storage and \pestel}
Documents are embedded (OpenAI TextEmbedding-3) and stored in Milvus with metadata: source, , publication date, \pestel\ label and version chain. A compact supervised classifier assigns a single \pestel\ label per item (multi-label is feasible but not used here). Cluster-level \pestel\ distributions are computed by aggregating item labels.

\section{Time-Dependent Recursive Summary Graph (TRSG)}
\label{sec:trsg}
\textbf{Goal.} Replace a flat feed with a weekly, two-level structure that is compact, faithful and easy to compare across weeks.

\paragraph{Cumulative Knowledge Base.}
While the crawling process runs daily and TRSG graphs are materialized weekly, all ingested data is stored in a single, persistent vector knowledge base. Each new crawl extends this base rather than overwriting it. This design allows the system to reason over historical embeddings when constructing new weekly hierarchies, enabling the detection of longer-term phenomena such as cluster drift, gradual topic convergence or the emergence of entirely new abstract groupings. Over time, these evolving structures provide early hints of potential future trends rather than merely weekly snapshots.

\begin{figure*}[t]
  \centering
  \includegraphics[width=\textwidth]{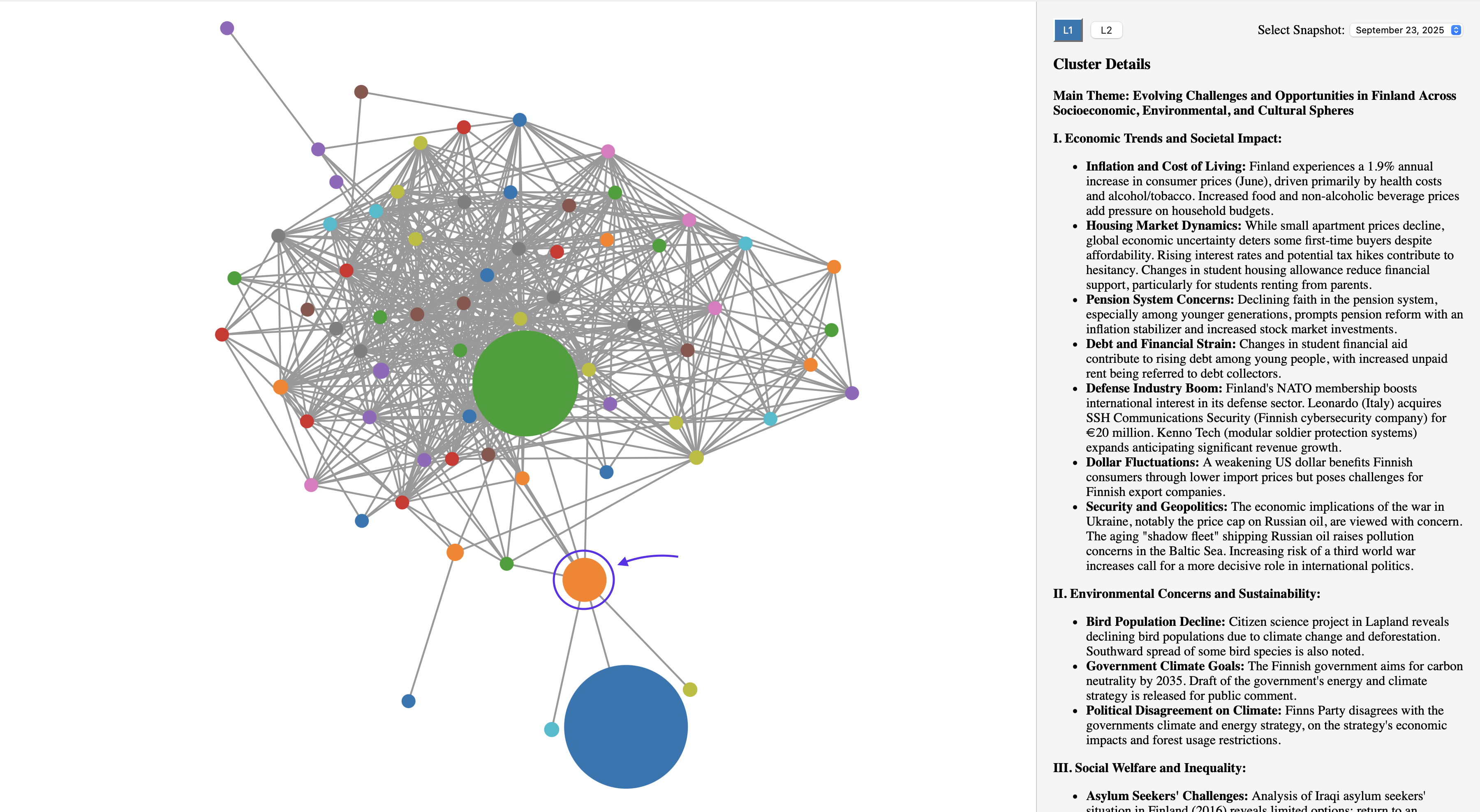}
  \caption{Example of the L1 layer in the TRSG. Each node represents a cluster of semantically related news items and edges indicate similarity links. The summarization step condenses each cluster into a factual thematic report, forming a connected graph of emerging narratives.}
  \label{fig:trsg_l1}
\end{figure*}

\subsection{Construction}
\paragraph{L0$\rightarrow$L1 (sub-clusters).} Build an item-level similarity graph for the week using cosine similarity; run Leiden \citep{traag2019louvain}. Summarize each community with a \emph{factual} prompt (names, dates, figures, relationships) and embed the resulting text as the L1 node.

\paragraph{L1$\rightarrow$L2 (meta-clusters).} Cluster L1 summaries and produce an \emph{abstract} summary per meta-cluster (themes, trends, implications, minimal specifics). L2 nodes represent the week’s landscape.

\paragraph{Stability knobs.} Small graphs use direct cosine; large graphs use FAISS with range thresholds. Weekly hierarchies are snapshotted for audit and fast reload.

\subsection{Prompt Design}
To enforce consistent abstraction, TRSG uses level-specific prompts and recursive summarization when input exceeds model limits.

\paragraph{L1 – Thematic summaries.}
Summarizes factual content across related news items:
\begin{quote}
\small
``Create a comprehensive L1 thematic summary in English. 
Do not evaluate or give suggestions. 
Identify the main theme, structure logically and include key facts—entities, dates, figures, policy changes and regional details. 
Write a complete, factual report without introductory remarks.''
\end{quote}
This produces grounded, information-rich cluster summaries.

\paragraph{L2 – Strategic synthesis.}
Aggregates L1 outputs into cross-domain insights:
\begin{quote}
\small
``Create a unified strategic L2 intelligence briefing in English. 
Do not evaluate or compare. 
Extract overarching patterns and systemic trends, emphasizing transformation forces across domains. 
Present as a coherent intelligence report without meta-commentary.''
\end{quote}
L2 abstracts factual clusters into foresight-level narratives.

\paragraph{Recursive summarization.}
If a cluster’s combined text exceeds the model’s context limit, the content is split into balanced chunks and summarized recursively. Each batch is first summarized individually with the same prompt and those interim summaries are then re-summarized to produce the final L1 or L2 output. This hierarchical compression maintains completeness and coherence even for large clusters.

\begin{table}[t]
\centering
\footnotesize
\setlength{\tabcolsep}{4pt}
\renewcommand{\arraystretch}{1.05}
\begin{tabularx}{\columnwidth}{l >{\raggedright\arraybackslash}X}
\toprule
\textbf{Component} & \textbf{Default / Behavior} \\
\midrule
Embedding model & \texttt{text-embedding-3-small} \\
L0$\rightarrow$L1 threshold & 0.75 (cosine) \\
L1$\rightarrow$L2 threshold & 0.55 (cosine) \\
Small-$n$ fallback & Direct cosine (no FAISS) \\
Clustering & Leiden (modularity) \\
Summarization & Gemini 2.0 Flash (L1 factual / L2 abstract) \\
Persistence & Weekly snapshots (\texttt{pickle}) \\
\bottomrule
\end{tabularx}
\caption{Key TRSG defaults used in production.}
\label{tab:defaults}
\end{table}

\section{Week-to-Week Change Detection}
\label{sec:week2week}
Snapshots from consecutive weeks are compared at L1 and L2 using cosine similarity.

\paragraph{Matching.} For each new summary, find the best old neighbour. Labels: \textbf{Stable} (sim $\ge$0.90), \textbf{Changed} (0.70–0.90), \textbf{Added} ($<$0.70). Unmatched old summaries are \textbf{Removed}. This yields structured deltas rather than vague impressions.

\paragraph{Theme grouping.} Added/Removed lists are converted to human-readable themes: short micro-labels per text (LLM), then canonicalization via TF--IDF + agglomerative clustering (cosine distance). The result is a deterministic set of \{\texttt{label}, \texttt{added\_texts}, \texttt{removed\_texts}\} per level.

\paragraph{\pestel\ analysis.} Users select perspectives. 
For each theme, a schema-constrained analysis returns 
\texttt{\{title, analysis, level, group, importance $\in$ [0,1]\}}. 
The results are cached in MySQL, keyed by week pair and perspective, 
to ensure reproducibility and fast retrieval.

\section{Use Case: Curriculum Intelligence}
An analyst monitoring Political+Technological developments compares weeks 23 and 28. TRSG highlights two new L2 themes: \emph{EU digital skills funding} and \emph{quantum computing policy momentum}. L1 reveals the concrete facts (program names, funding figures, named institutions). The \pestel\ analysis recommends actions: (i) align elective modules with EU skill frameworks, (ii) add a quantum fundamentals stack (concepts + labs) and (iii) explore partnerships with local industry labs. The value is not prediction but \emph{traceable synthesis} tailored to University’s remit.

In addition to supporting concrete curriculum adjustments, the TRSG output provides a reusable evidence base for cross-faculty coordination. Because every proposed action links back to the underlying news signals and cluster summaries, departments can justify decisions using a shared source of truth rather than ad-hoc interpretations or anecdotal reports. This auditability also supports institutional learning: when decisions are revisited months later, the exact informational context that motivated them remains inspectable. In practice, the platform enables routine, low-friction foresight workflows (monthly horizon scans, annual strategy cycles or accreditation preparations) without requiring analysts to rebuild situational awareness from scratch.

The same machinery also generalizes to other decision layers, including research prioritization, stakeholder engagement and regional partnership planning. Since TRSG captures both stable background narratives and emerging weak signals, it helps distinguish between noise, structural change and transient bursts of attention. For universities operating in rapidly evolving technological and policy environments, this distinction is essential: long-horizon initiatives (e.g., lab infrastructure, degree redesign) demand evidence of durable trends, while quick interventions (e.g., micro-credential pilots) benefit from timely detection of nascent opportunities. By encoding both views into a single recurring graph, \system\ makes such multiscale reasoning feasible for non-technical users.

\section{Conclusion}
\system\ demonstrates that continuous, traceable foresight over fast-moving news streams is operationally achievable using a combination of embeddings, hierarchical clustering, recursive summarization and week-to-week change detection. By structuring evolving narratives into a two-level \ours\ and grounding interpretation in \pestel, the platform delivers decision-ready intelligence that remains auditable and aligned with institutional needs. The curriculum-intelligence case shows that such a system can inform real strategic choices without relying on opaque prediction. Future work will refine evaluation, explore multi-label \pestel, and extend the approach to multilingual sources and policy–science–industry link analysis.

\section{Limitations}
Our system has several inherent limitations:

\begin{itemize}[leftmargin=1.1em,itemsep=1pt,topsep=2pt]
  \item \textbf{Coverage bias.} The system depends on which news sources are crawled; underrepresented voices or niche media may be missed, skewing the narrative graph.
  \item \textbf{Summary hallucination.} Although prompts are guarded and ground traces are preserved, LLMs may still introduce inaccuracies or omit subtle but relevant facts.
  \item \textbf{Domain specificity.} The \pestel\ classifier and thresholds are tuned for one university’s domain; generalizing to another institution or domain will require re-training or re-tuning.
  \item \textbf{Temporal granularity.} Weekly snapshots may miss rapid developments or sub-week bursts; while daily crawling accumulates data, changes within a week are abstracted.
\end{itemize}

\bibliography{custom}
\end{document}